\documentclass[final,OA]{AAAI-Std}
\usepackage{atbegshi} 
\makeatletter
\def\FigPath{.}
\def\opsfilepath{.}
\renewcommand{\graphicspath}{{./}}
\makeatother

\usepackage{array}
\usepackage{makecell}
\usepackage{tabularx}
\usepackage{ragged2e}
\usepackage{tablefootnote} 
\usepackage{threeparttable}
\usepackage{booktabs} 
\usepackage{graphicx}
\usepackage{caption}
\usepackage{subcaption}
\usepackage{float}
\usepackage{listings}
\usepackage{xcolor}

\definecolor{white}{rgb}{0, 0, 0}
\newcommand{\majorrevision}[1]{\textcolor{white}{{#1}}}

\begin{document}

\title{S\textsc{truct}S\textsc{ense}: A T\textsc{ask-agnostic} A\textsc{gentic} F\textsc{ramework} for
       S\textsc{tructured} I\textsc{nformation} E\textsc{xtraction} with
       H\textsc{uman-in-the-loop} E\textsc{valuation} and B\textsc{enchmarking}}

\author[1]{T\textsc{ek} R\textsc{aj} C\textsc{hhetri}}
\author[1]{Y\textsc{ibei} C\textsc{hen}} 
\author[1]{P\textsc{uja} T\textsc{rivedi}} 
\author[1]{D\textsc{orota} J\textsc{arecka}} 
\author[2]{S\textsc{aif} H\textsc{aobsh}} 
\author[3]{P\textsc{atrick} R\textsc{ay}} 
\author[3]{L\textsc{ydia} N\textsc{g}} 
\author[1]{S\textsc{atrajit} S. G\textsc{hosh}} 

\address[1]{McGovern Institute for Brain Research, Massachusetts Institute of Technology, Cambridge, MA, USA}
\address[2]{Fylo Labs Inc., New York, NY, USA}
\address[3]{Allen Institute for Brain Science, Seattle, WA, USA}

\abstract[Abstract]{
Extracting structured information from scientific literature is critical 
for accelerating discovery, yet Large Language Models (LLMs) often struggle in specialized domains that require expert knowledge and generalize poorly across tasks. We introduce \textsc{StructSense}, a modular, 
task-agnostic, open-source framework that integrates ontology-guided 
symbolic knowledge, agentic self-evaluative refinement, and 
human-in-the-loop validation for robust domain-aware extraction. We 
evaluate \textsc{StructSense} on three tasks of increasing semantic 
complexity: schema-based extraction of assessment instruments 
(91--100\% accuracy), metadata and resource extraction from scientific 
papers (86--93\% overall), and named entity recognition (NER) from 
neuroscience literature (58--75\% label accuracy across 8,882 
entities). On two biomedical NER benchmarks (NCBI Disease and S800 
Species), the system achieves $\geq$90\% relaxed recall and 
62.5--85.8\% strict recall while extracting 1,000--3,600 additional 
entities beyond gold annotations. The local concept mapping service 
achieves Hits@1 of 62--82\% under strict matching and 68--86\% under 
semantic matching. These results across three domains demonstrate that 
\textsc{StructSense} generalizes across tasks while maintaining 
source grounding and provenance transparency.}


\maketitle

\section{Introduction}\label{sec:intro}
The latest National Science Foundation report estimated 3.3 million new scientific articles published in 2022~\citep{nsb2023publications}, up from 2.2 million in 2016 ~\citep{Nasar2018}. This growth is projected to accelerate: in March 2026 alone, arXiv\footnote{\url{https://arxiv.org}} received 30,046 new submissions\footnote{\url{https://arxiv.org/stats/monthly_submissions}} and bioRxiv\footnote{\url{https://www.biorxiv.org}} recorded 1,180 neuroscience manuscripts\footnote{Count manually retrieved by the authors for the neuroscience category on bioRxiv.}. This rapid expansion has overwhelmed researchers' capacity to track advances, evaluate quality, and integrate emerging findings~\citep{Landhuis2016, Nasar2018, li2021scientificdiscoursetaggingevidence}.

This challenge has driven growing interest in \textit{structured information extraction~(SIE)}, also called information extraction~(IE)---terms we use interchangeably---for curating scientific knowledge~\citep{Verma2023, Dagdelen2024, Xu2024}. It has also spurred open science initiatives such as the Fylo\footnote{\url{https://www.fylo.io}} project, an Astera-supported effort focused on SIE from scientific literature. SIE converts unstructured text into structured representations such as entities, events, and relationships, and underpins knowledge graph construction, semantic search, and scientific question answering~\citep{Xu2024, 10.1145/3701716.3715483, Hong2021}.

IE systems have traditionally performed named entity recognition (NER) followed by relation extraction (RE) in multi-stage pipelines~\citep{Dagdelen2024}. These modular architectures, while effective, often suffer from cascading errors. Recent LLM~(Large Language Model) advances have shifted this paradigm toward end-to-end IE with fewer architectural constraints~\citep{Schröder2022, agrawal2022largelanguagemodelsfewshot, Dagdelen2024, Gupta2024}, enabling more flexible and integrated frameworks.

Despite this progress, several challenges persist in applying IE to scientific literature: (i)~limited understanding of domain-specific terminology and nonstandard phrasing; (ii)~difficulty handling polysemous terms (e.g., ``cortex'')~\citep{agrawal2022largelanguagemodelsfewshot, wan2023gptreincontextlearningrelation}; (iii)~inadequate provenance tracking for extracted facts; and (iv)~lack of transparency and reproducibility. Most current systems also fail to produce FAIR~(Findable, Accessible, Interoperable, and Reusable)-compliant outputs~\citep{wilkinson2016fair} or meet criteria for AI readiness and reproducibility~\citep{10.1145/3701716.3715483, Kirkpatrick2023}. Ontologies---formal specifications of shared domain concepts and their relationships~\citep{Fensel2004}---can address several of these challenges by providing standardized vocabularies and semantic grounding for extraction. Schemas similarly represent domain conceptualizations but lack the expressive power to model the complex relationships and logical constraints that ontologies capture.

In the biomedical domain, NER benefits from rich annotated corpora (e.g., BC5CDR~\citep{li2016bc5cdr}, BioNLP~\citep{kim2009overview}, JNLPBA~\citep{collier2004introduction}) and pre-trained models such as BioBERT~\citep{lee2020biobert} and PubMedBERT~\citep{gu2022domain}. Neuroscience, however, lacks comparable annotated datasets and ontological resources, and its diverse constructs---neural systems, cognitive processes, experimental paradigms---are poorly covered by existing biomedical NER tools~\citep{pyysalo2013distributional, cho2021neural}.

We introduce \textsc{StructSense}, a modular, task-agnostic, open-source multi-agent system that integrates curated domain knowledge via ontologies, replaces brittle pipelines with an agentic architecture featuring self-evaluative refinement, and incorporates human-in-the-loop validation. The system structures extracted entities for direct ingestion into ML workflows, and its tool-use and memory modules support reproducible extraction across experimental runs. 

By combining LLM-based information extraction with ontology-driven concept mapping, \textsc{StructSense} transforms unstructured domain content into semantically normalized, structured outputs. This design supports key FAIR  principles by improving findability through standardized semantic annotations and by promoting interoperability and reusability through the reuse of established ontologies. Moreover, by partially automating the structuring and normalization of domain-specific data, \textsc{StructSense} contributes to making such outputs more AI-ready and suitable for reliable downstream use. 

\majorrevision{We demonstrate \textsc{StructSense} across three case studies. Case Study~I evaluates three neuroscience-domain tasks: (1)~NER from full-text articles\footnote{\url{https://github.com/sensein/structsense/tree/main/evaluation/ner}}, (2)~conversion of assessment instruments into ReproSchema~\citep{Chen2025ReproSchema}\footnote{\url{https://github.com/sensein/structsense/tree/main/evaluation/pdf2reproschema}}, and (3)~extraction of scientific resources and metadata\footnote{\url{https://github.com/sensein/structsense/tree/main/evaluation/resource_extraction}}. Case Study~II benchmarks biomedical NER on two standard datasets (NCBI Disease and S800 Species). Case Study~III independently evaluates the local concept mapping service under multiple retrieval and re-ranking configurations. Together, these case studies span diverse domains, formats, and extraction goals, illustrating cross-task and cross-domain generalizability.}

\section{Related Works}\label{sec:related-works}

\majorrevision{We organize related work into four categories: rule-based and ontology-guided approaches, ML/DL-based approaches, LLM-based approaches, and agentic or tool-augmented systems. We treat LLM-based methods separately from conventional ML/DL due to the paradigm shift they introduced in language processing.}

\subsection{Rule-based and Ontology-guided IE}
\label{sec:sota_rule_based}

\majorrevision{Rule-based IE systems leverage handcrafted patterns and domain ontologies for extraction. ~\citep{9366868} combined regular expressions with fuzzy rules and Word2Vec for content and metadata extraction from scientific documents, but their narrow ontology scope and rigid rules limit generalizability to diverse document formats. ~\citep{Schröder2022} applied structured rules to extract provenance information from wet-lab protocols, enabling complex SPARQL queries; however, their approach cannot adapt to unseen protocol structures or linguistic variability. ~\citep{doi:10.1177/0165551515610989} extended ontology-based IE with hybrid extraction (regular expressions + CRF) and ontology-based error detection, supporting flexible integration of third-party ontologies from BioPortal. Their evaluation on synthetic student-response data, however, limits claims about real-world generalizability. ~\citep{LIU2017313} combined ontology-based IE with semi-supervised CRF to reduce dependence on labeled data, but acknowledged limited ability to extract relations between entities.}

\majorrevision{A common thread across these approaches is strong domain semantics and alignment with FAIR principles, offset by brittleness to linguistic variability and limited adaptability to new domains without manual rule engineering.}

\subsection{ML and Deep Learning Approaches}
\label{sec:sota_ml_based}

\majorrevision{ML and DL approaches learn extraction patterns from data, reducing reliance on handcrafted rules. ~\citep{10101029} proposed a CNN--BiLSTM ensemble for scholarly metadata extraction that outperformed decision trees and SVMs, but relied heavily on layout cues (e.g., font-based features), limiting robustness across document formats. Andruszkiewicz and Rybinski~\citep{8334468} combined CRFs and SVMs with rule-based techniques to extract publication metadata and link them to researchers, though their approach considered only a narrow set of entity types. Both approaches require domain-specific training data, reducing adaptability to new domains without re-annotation and re-training.}

\subsection{LLM-based IE}\label{sec:sota_llm_based}

\majorrevision{LLMs enable flexible, few-shot IE without task-specific architectures. ~\citep{agrawal2022largelanguagemodelsfewshot} demonstrated few-shot learning for clinical IE tasks including abbreviation disambiguation, coreference resolution, and medication extraction, but noted that LLMs struggle to adhere to exact output schemas without external tool support. ~\citep{li2021scientificdiscoursetaggingevidence} achieved state-of-the-art scientific discourse tagging on two benchmarks and demonstrated utility for claim extraction; their approach, however, does not ensure alignment with curated domain ontologies. ~\citep{wei2023claimdistiller} framed claim extraction as binary sentence classification. Similarly, other studies~\citep{huguet-cabot-navigli-2021-rebel-relation, Wu02012023} extract individual components---either entities or relations---rather than complete structured outputs, requiring multi-stage pipelines that reduce efficiency~\citep{Dagdelen2024}.}

\majorrevision{These LLM-based approaches demonstrate strong language understanding but typically lack integration with domain knowledge, limiting accuracy in specialized fields. They also exhibit limited task generality without fine-tuning~\citep{10.1007/978-3-031-77844-5_7}.}

\subsection{Agentic and Tool-augmented IE}
\label{sec:sota_agentic}

\majorrevision{Recent work has explored agentic architectures in which LLMs orchestrate tool use, self-evaluation, and iterative refinement for complex tasks~\citep{gottweis2025aicoscientist, wu2024metarewardinglanguagemodelsselfimproving}. However, the application of agentic frameworks to structured information extraction remains largely unexplored. Existing IE systems generally do not leverage tool-augmented reasoning, self-evaluative feedback loops, or human-in-the-loop collaboration within an agentic workflow. This gap motivates the design of \textsc{StructSense}.}

\subsection{Summary and Positioning}\label{sec:sotasummary}

\majorrevision{Rule-based methods offer strong domain semantics but lack adaptability; ML/DL approaches learn from data but require large annotated datasets; LLM-based systems provide flexible language understanding but lack domain knowledge integration and task generality; and agentic IE remains nascent. \textsc{StructSense} addresses these gaps by combining ontology-guided knowledge grounding with an agentic multi-agent architecture, incorporating self-evaluative refinement, human-in-the-loop validation, explicit source grounding, and provenance tracking---while operating in a task-agnostic manner without fine-tuning.}

\section{\textsc{StructSense}}\label{sec:framework}

\begin{figure*}[t]
    \centering
    \includegraphics[width=1\linewidth]{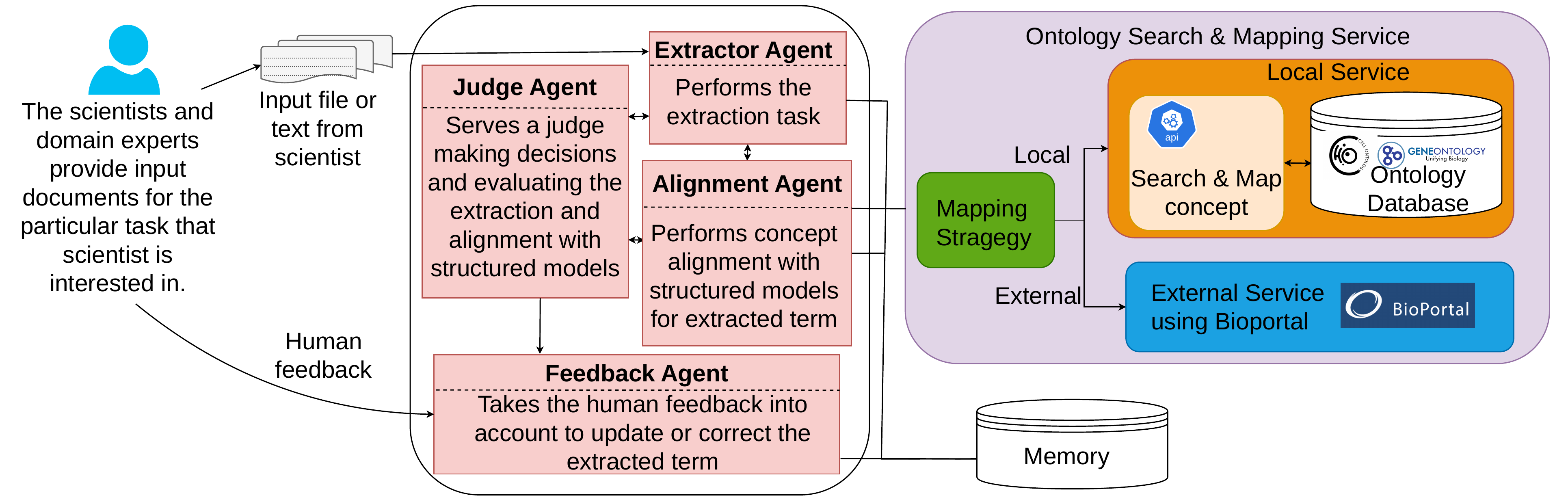}
    \caption{Architecture of \textsc{StructSense}.}
    \label{fig:structsensearch}
\end{figure*}

\majorrevision{Figure~\ref{fig:structsensearch} shows the architecture of \textsc{StructSense}, a task-agnostic multi-agent framework that orchestrates the extraction, evaluation, and refinement of structured data from unstructured sources. The system decomposes into four specialized agents: (i)~extractor, (ii)~alignment, (iii)~judge, and (iv)~feedback. This modular design improves reusability, simplifies maintenance, and facilitates the management of complex extraction workflows. Each agent receives two configurations as input: an \textit{agent configuration} specifying the LLM and API keys, and a \textit{task configuration} defining objectives, input/output formats, and extraction constraints.}

\subsection{Extractor Agent}\label{sec:extractoragent}

\majorrevision{The extractor agent extract the information from unstructured text and produce a structured JSON output whose content and structure vary by task configuration. Unlike existing works\footnote{\url{https://edisonscientific.gitbook.io/edison-cookbook/edison-client}} that rely solely on LLMs, the extractor leverages specialized, domain-specific tools invoked automatically based on the task type. Task detection relies primarily on LLM-based inference, with keyword-driven heuristics as a fallback.}

\begin{figure*}[t]
    \centering
    \includegraphics[width=1\linewidth]{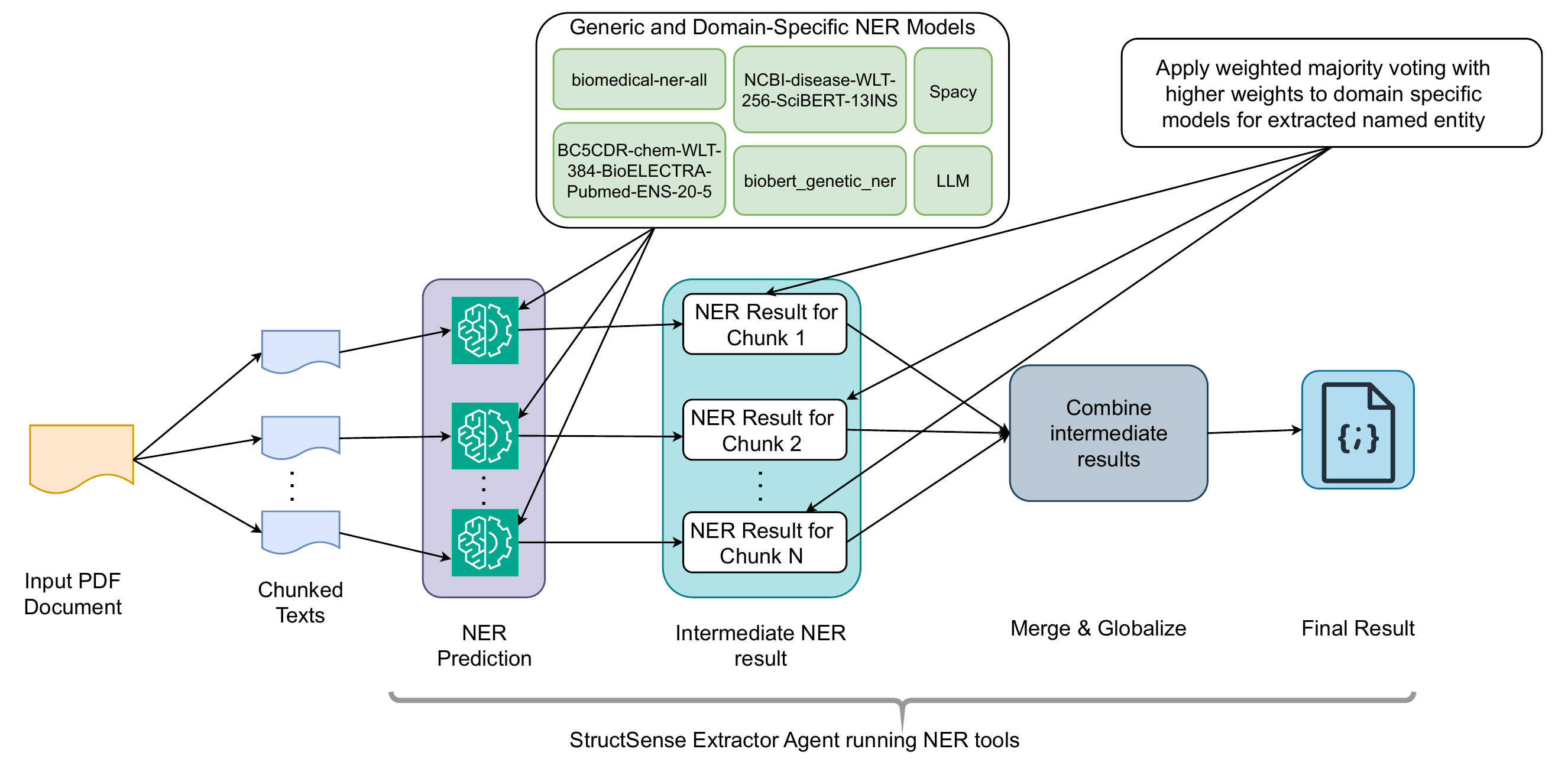}
    \caption{Overview of the extractor agent architecture, 
    illustrating the use of specialized tools and 
    domain-specific NER models along with an LLM for named 
    entity extraction under chunking-based processing.}
    \label{fig:nertoolcallextraction}
\end{figure*}

\majorrevision{Figure~\ref{fig:nertoolcallextraction} illustrates the NER extraction workflow in chunking mode. The input text is segmented into chunks based on a user-defined chunk size. For each chunk, NER is performed using an LLM together with a combination of domain-specific models and a general-purpose SpaCy-based model. The specific models used are detailed in \S\ref{sec:structsenseimplementation}. Chunk-level outputs are then consolidated into a unified global entity set during a merge and globalization\footnote{ Globalization means that all entities from the chunked/parallel runs are mapped back to the full text, providing global spans. Here, full text refers to the text extracted by Grobid that is passed as input to the extractor agent.} phase.}

\majorrevision{During merging, a multi-level verification process ensures strict grounding in the source text, mitigating LLM hallucinations. First, if an entity contains valid global span offsets, the corresponding substring is compared against the normalized entity text; if span validation fails, a fallback checks whether the entity text appears anywhere in the source (case-insensitive). Entities failing both checks are discarded. Second, for entities with valid spans, sentence-level verification confirms that each recorded source sentence exists in the original text; entities with ungrounded sentences are removed. This verification applies across all tasks---for example, resource extraction verifies that extracted datasets, models, and URLs are grounded in the source. The modular architecture enables seamless extension with additional domain-specific models, tools, or verification mechanisms. For NER, we use a weighted ensemble with the following model weights: (i) \textit{biomedical-ner-all}, 5.0; (ii) \textit{biobert\_genetic\_ner}, 4.0; (iii) \textit{llm\_ner}, 3.9; (iv) \texttt{BC5CDR-chem-\allowbreak WLT-384-\allowbreak BioELECTRA-\allowbreak Pubmed-\allowbreak ENS-20-5}, 3.0; (v) \textit{NCBI-disease-WLT-256-SciBERT-13INS}, 2.0; and (vi) \textit{en\_core\_web\_sm}, 1.0. This weighting scheme intentionally prioritizes domain-specific models over the LLM. The sensitivity of the overall system to these weight assignments, and their impact on extraction performance, remain important directions for future investigation.}

\subsection{Alignment Agent}\label{sec:alignmentagent}

\majorrevision{The alignment agent maps extracted entities to canonical concepts in ontologies or knowledge graphs. This step provides the curated domain knowledge that LLMs lack~\citep{jiang-etal-2024-efficient, wang-etal-2024-infuserki, xie-etal-2022-unifiedskg} and resolves ambiguities arising from polysemous terms. For example, as shown in Table~\ref{tab:cortex-ambiguity}, ``cortex'' denotes a brain region in neuroscience (UBERON\_0000956) but an outer cell layer in plant biology (PO\_0005708). Without ontological grounding, an LLM may select the wrong sense. The alignment agent links each occurrence to the appropriate ontology term, ensuring context-aware outputs.}

\begin{table*}[t]
\small
\centering
\begin{threeparttable}
\caption{Domain-specific semantics of the polysemous term 
\emph{cortex}: contrasting anatomical structures in 
neuroscience versus plant biology.}
\label{tab:cortex-ambiguity}
\begin{tabularx}{\textwidth}{@{}p{1.4cm}p{2.5cm}X p{2.9cm}X@{}}
\toprule
\textbf{Term} &
\textbf{Neuroscience-related ontology ID} &
\textbf{Neuroscience meaning} &
\textbf{Ontology ID in plant biology domain} &
\textbf{Contrasting meaning in plant biology} \\[2pt]
\midrule
\emph{cortex} &
UBERON\_0000956\tnote{a} &
The outermost region of the brain represents the anatomical 
information about the brain. &
PO\_0005708\tnote{b} &
A primary plant tissue located between the epidermis and the 
central vascular cylinder, primarily involved in storage, 
transport, and structural support. \\
\bottomrule
\end{tabularx}
\begin{tablenotes}
\small
\item[a]\url{http://purl.obolibrary.org/obo/UBERON_0001851}
\item[b]\url{http://purl.obolibrary.org/obo/PO_0005708} 
\end{tablenotes}
\end{threeparttable}
\end{table*}

\majorrevision{Additionally, by accessing external domain knowledge, the alignment agent minimizes the need to pretrain or fine-tune LLMs for every domain-specific knowledge update, avoiding the inefficiencies of models relearning existing knowledge~\citep{wang-etal-2024-infuserki}. The agent delegates concept mapping to a configurable service that supports both a local concept mapping backend and BioPortal~\citep{10.1093/nar/gkaf402} for external ontology-based mapping. The local service is described in detail in \S\ref{sec:localconceptmappingservice}.}

\subsection{Judge Agent}\label{sec:judgeagent}

\majorrevision{The judge agent serves as a critic that evaluates the alignment agent's output---specifically, the extracted structured information and its corresponding ontological mappings. Its design follows prior work~\citep{10572486, wu2024metarewardinglanguagemodelsselfimproving, gottweis2025aicoscientist} demonstrating that self-evaluation mechanisms in agent workflows improve overall performance. The judge assigns a confidence score from 0 to 1 (where 1 denotes a perfect match), following established benchmarks that adopt similar LLM-based scoring methodologies~\citep{NEURIPS2023_91f18a12}. In addition to the score, the judge records remarks that provide the rationale underlying each assessment, supporting interpretability.}

\subsection{Feedback Agent}\label{sec:feedbackagent}

\majorrevision{The feedback agent integrates a human-in-the-loop (HIL) component. Despite their capabilities, LLMs remain prone to hallucinations---producing content that appears plausible but lacks factual grounding~\citep{10.1145/3703155}. Human oversight enhances reliability and trustworthiness by mitigating such errors~\citep{li2025largelanguagemodelsstruggle, amirizaniani2024llmauditorframeworkauditinglarge}. The feedback agent allows users to correct outputs or provide guidance in natural language before the final result is generated.}

\subsection{Memory}\label{sec:memory}

\majorrevision{Memory enables agents to maintain contextual awareness throughout the extraction pipeline. \textsc{StructSense} supports three memory types: contextual memory, entity memory, and long-term memory. These modules are shared across agents, allowing them to retain relevant context, previously identified entities, and knowledge acquired during execution. Shared memory supports effective collaboration and contextual continuity across the pipeline.}

\subsection{Provenance}\label{sec:provenance}

\majorrevision{A key factor in establishing trust is understanding where and how predictions originate. \textsc{StructSense} treats provenance as a first-class design principle, differentiating it from existing works. During extraction, the system captures fine-grained provenance metadata for each prediction, including the contributing models (Table~\ref{tab:ner_detectionprovenancemodel}). During concept mapping, it records whether each output was produced by the mapping tool or inferred from the LLM's parametric knowledge (Table~\ref{tab:ner_ontologyprovenance}). The judge agent stores the remarks underlying each assigned score (Table~\ref{tab:ner_llmjudgescoreremarksprovenance}). This explicit attribution improves traceability and can improve overall trust.}

\begin{table*}[t]
\centering
\caption{NER output showing extracted entities and associated 
provenance metadata. \textbf{Extracted sentence:} 
\textit{``Ataxia - telangiectasia (A-T) is a recessive 
multi-system disorder caused by mutations in the ATM gene at 
11q22--q23.''}}
\label{tab:ner_detectionprovenancemodel}
\renewcommand{\arraystretch}{1.35}
\resizebox{\textwidth}{!}{%
\begin{tabular}{p{4cm} l c c r r r r}
\toprule
\textbf{Entity} & \textbf{Label} & \textbf{Global} & 
\textbf{Local} & \textbf{Weighted} & \textbf{Model} & 
\textbf{Source} & \textbf{Vote} \\
 & & \textbf{Span} & \textbf{Span} & \textbf{Score} & 
\textbf{Count} & \textbf{Model} & \textbf{Weight} \\
\midrule
Ataxia-telangiectasia & DISEASE & 104--127 & 0--23 & 1.0 & 1 
& \texttt{llm\_ner} & 3.9 \\
ATM & GENE & 204--207 & 100--103 & 1.0 & 1 & 
\texttt{llm\_ner} & 3.9 \\
\bottomrule
\end{tabular}%
}
\smallskip
\begin{minipage}{\textwidth}
\footnotesize
\textit{Global span}: character offsets in the full document. \textit{Local span}: character offsets within the containing sentence. \textit{Weighted score}: ensemble-aggregated confidence across all source models. \textit{Vote weight}: weight assigned to each model for aggregation via weighted majority voting. \textit{Source model}: name of the extraction model; \textit{llm\_ner} denotes LLM-based inference. \textit{Model count}: total number of participating models; here the output is generated by the LLM only (count = 1).
\end{minipage}
\end{table*}

\begin{table*}[t]
\centering
\caption{Concept mapping provenance for extracted entities. \textbf{Extracted sentence:} \textit{``Ataxia - telangiectasia (A-T) is a recessive multi-system disorder caused by mutations in the ATM gene at 11q22--q23.''}}
\label{tab:ner_ontologyprovenance}
\renewcommand{\arraystretch}{1.35}
\resizebox{\textwidth}{!}{%
\begin{tabular}{l l l l c}
\toprule
\textbf{Entity} & \textbf{Ontology} & \textbf{Ontology Label} 
& \textbf{Ontology ID (or IRI)} & \textbf{Prov.} \\
\midrule
\multirow{2}{*}{Ataxia-telangiectasia}
  & NDFRT & Ataxia Telangiectasia {[}Disease/Finding{]}
  & \href{http://purl.bioontology.org/ontology/NDFRT/N0000000503}{\texttt{http://purl.bioontology.org/ontology/NDFRT/N0000000503}}
  & tool \\ 
  & IOBC & Osgood-Schlatter disease
  & \href{http://purl.jp/bio/4/id/200906063051059008}{\texttt{http://purl.jp/bio/4/id/200906063051059008}}
  & tool \\
\midrule
\multirow{2}{*}{ATM} 
  & IOBC & ATM gene
  & \href{http://purl.jp/bio/4/id/200906077355454486}{\texttt{http://purl.jp/bio/4/id/200906077355454486}}
  & tool \\ 
  & IOBC & KiSS1 gene
  & \href{http://purl.jp/bio/4/id/200906004294523822}{\texttt{http://purl.jp/bio/4/id/200906004294523822}}
  & tool \\
\bottomrule
\end{tabular}%
}
\smallskip
\begin{minipage}{\textwidth}
\footnotesize 
Top-2 concept mapping results with provenance metadata. \textbf{Prov.}: indicates how the mapping was obtained; \textit{tool} indicates the output is from the concept mapping service rather than the LLM's parametric knowledge. \textbf{Ontology}: ontology acronym. \textbf{Ontology Label}: label of the mapped concept. \textbf{Ontology ID} (or \textit{IRI}): class IRI of the mapped concept.
\end{minipage}
\end{table*}

\begin{table*}[t]
\centering
\caption{LLM-as-a-Judge provenance, capturing the rationale (remarks) underlying the assigned score. \textbf{Extracted sentence:} \textit{``Ataxia - telangiectasia (A-T) is a recessive multi-system disorder caused by mutations in the ATM gene at 11q22--q23.''}}
\label{tab:ner_llmjudgescoreremarksprovenance}
\renewcommand{\arraystretch}{1.35}
\begin{tabular}{l l c l}
\toprule
\textbf{Entity} & \textbf{Label} & \textbf{Judge Score} & 
\textbf{Remarks} \\
\midrule
Ataxia-telangiectasia & DISEASE & 1.0 & Perfect identification 
and mapping. \\
ATM & GENE & 1.0 & Accurately identified as a gene. \\
\bottomrule
\end{tabular} 
\end{table*}

\section{Local Concept Mapping Service}
\label{sec:localconceptmappingservice}

\begin{figure*}[t]
    \centering
    \includegraphics[width=0.8\linewidth]{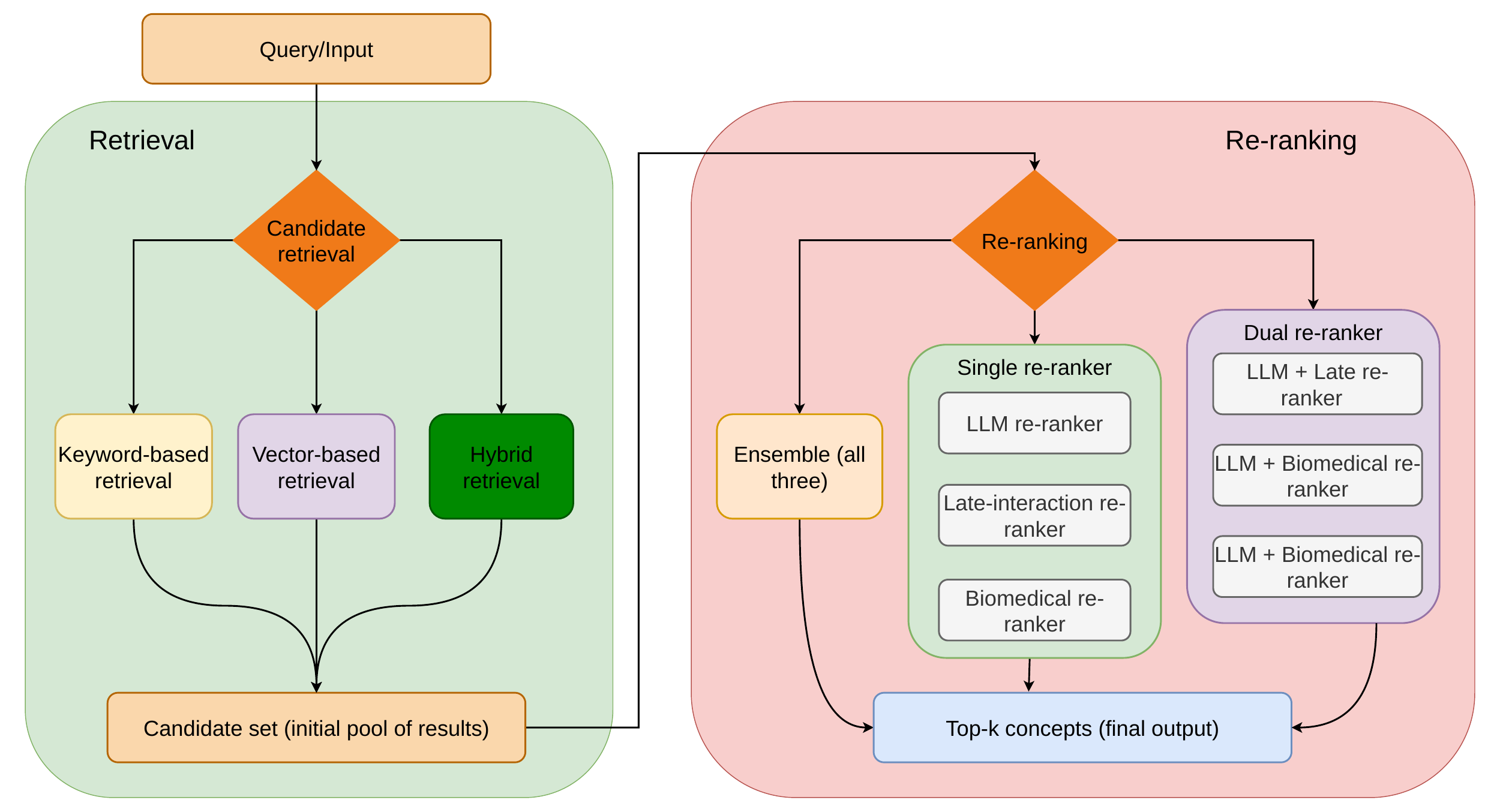}
    \caption{Core pipeline of the ontology search and mapping 
    service, showing retrieval and re-ranking components that 
    produce the final top-$k$ concept candidates for a given 
    query.}
    \label{fig:connceptmappingtool}
\end{figure*}

\majorrevision{We formulate concept mapping as an information retrieval task: given a query $q$ comprising an extracted entity and its textual context, the goal is to identify the best-matching ontological concept $d$. For example, the extracted entity \emph{Ataxia Telangiectasia} maps to the top-ranked concept \emph{Ataxia Telangiectasia [Disease/Finding]} from the NDFRT ontology (Table~\ref{tab:ner_ontologyprovenance}). Figure~\ref{fig:connceptmappingtool} illustrates the core pipeline, which comprises three stages: (i)~index and embedding generation, (ii)~retrieval, and (iii)~re-ranking.}

\paragraph{Index and embedding generation} For each ontology concept $d$, we construct a rich-text representation $D(d)$ by concatenating all available fields:

\begin{equation}
D(d) = \bigl[l_{\text{pref}} \oplus t_{\text{def}} \oplus 
L_{\text{alt}} \oplus Syn \oplus Par \oplus O_{\text{id}}\bigr]
\label{eq:richtext}
\end{equation}

\noindent \majorrevision{where $\oplus$ denotes whitespace-separated concatenation  of non-empty fields. Table~\ref{tab:conceptequation} describes each symbol. The ontology identifier $O_{\text{id}}$ is appended in textual form (e.g., \texttt{ontology: NIFSTD}) so that the ontology name contributes only a small signal during retrieval.}

\majorrevision{This representation $D(d)$ serves as input to both sparse indexing (BM25) and dense embedding generation. Dense vectors are produced using the \texttt{bge-small-en-v1.5} model~\citep{bge_embedding}. The service supports multiple retrieval backends; specific backend configurations are described in \S\ref{sec:localconceptmapping}.}

\begin{table}[t]
\centering
\begin{tabular}{lp{5.2cm}}
  \toprule
  Symbol & Description \\
  \midrule
  $l_{\text{pref}}$ & Preferred label of concept $d$. \\
  $t_{\text{def}}$ & Definition of concept $d$. \\
  $L_{\text{alt}}$ & Alternative labels of concept $d$. \\
  $Syn$ & Synonyms of concept $d$. \\
  $Par$ & Parent concept labels in the ontology hierarchy. \\
  $O_{\text{id}}$ & Ontology acronym (e.g., \texttt{NIFSTD}). \\
  \bottomrule
\end{tabular}
\caption{Symbols used in the rich-text representation 
(Equation~\ref{eq:richtext}).}
\label{tab:conceptequation}
\end{table}

\majorrevision{\paragraph{Retrieval} The system constructs a retrieval query $q_{\text{retrieve}}$ by appending optional context $c$ to the entity mention $q$:}

\begin{equation}
  q_{\text{retrieve}} =
  \begin{cases}
    q \mathbin{+} \texttt{" "} \mathbin{+} c 
      & \text{if context is available} \\
    q & \text{otherwise}
  \end{cases}
  \label{eq:qretrieve}
\end{equation}

\majorrevision{Three retrieval strategies are supported. \textit{Keyword-based retrieval} tokenizes $q_{\text{retrieve}}$ and scores candidates using BM25 weighting with saturated term frequency~\citep{lu2024bm25sordersmagnitudefaster}. \textit{Vector-based retrieval} encodes $q_{\text{retrieve}}$ into a dense vector and retrieves candidates by cosine similarity against pre-computed embeddings. \textit{Hybrid retrieval} combines both methods using a weighted scheme that controls each method's contribution. All three strategies return an initial top-$k$ candidate set.}

\paragraph{Re-ranking}
\majorrevision{Re-ranking reorders the retrieved candidates to produce the final top-$k$ results. The service supports three re-rankers: (i)~an \textit{LLM-based} re-ranker, (ii)~a \textit{late-interaction} re-ranker that encodes queries and documents independently and delays their interaction until scoring~\citep{10.1145/3397271.3401075}, and (iii)~a \textit{biomedical keyword} re-ranker. The late-interaction and biomedical re-rankers operate fully offline. Beyond individual use, the framework supports dual combinations of any two re-rankers and ensemble combinations of all three. All re-ranking is pointwise: each candidate is scored independently. For dual or ensemble configurations, scores are combined using a normalized weighted scheme. Specific model choices and weight configurations are described in \S\ref{sec:localconceptmapping}.}

\section{Implementation}\label{sec:implementation}

The source code for \textsc{StructSense} and usage examples are available at \url{https://github.com/sensein/structsense}. Indexes and embeddings for the local concept mapping service are hosted on Hugging Face at \url{https://huggingface.co/datasets/sensein/ontology-sqlite-vectorstore}, and its source code is available at \url{https://github.com/sensein/search_hybrid}.

\subsection{StructSense}\label{sec:structsenseimplementation}

\majorrevision{\textsc{StructSense} is implemented in Python~3 using the Crew.AI\footnote{\url{https://docs.crewai.com/en}} framework for multi-agent orchestration. For document parsing, it uses GROBID\footnote{\url{https://grobid.readthedocs.io/en/latest}}, an ML library for extracting structured information from scholarly documents, wrapped by GrobidArticleExtractor\footnote{\url{https://github.com/sensein/GrobidArticleExtractor}}, which organizes extracted content by document section. This sectioned text serves as input to the agents.}

\majorrevision{\paragraph{Processing modes} The system supports parallel agent execution in both chunked and non-chunked modes. In chunked mode, the input text is parsed using SpaCy, partitioned into segments~(default: 2,000 characters), and distributed to agents for parallel processing. Chunk-level outputs are aggregated into a consolidated result. Verification is performed after extraction and again at the end of the pipeline; intermediate stages~(e.g., alignment) are excluded from verification because they extend outputs with additional concept-level information.}

\majorrevision{\paragraph{NER models} For the NER task, the extractor agent invokes an LLM together with five domain-specific models: a biomedical entity model\footnote{\url{https://huggingface.co/d4data/biomedical-ner-all}}, a disease model trained on NCBI\footnote{\url{https://huggingface.co/mobashgr/NCBI-disease-WLT-256-SciBERT-13INS}}, a chemical entity model based on BC5CDR\footnote{\url{https://huggingface.co/mobashgr/BC5CDR-chem-WLT-384-BioELECTRA-Pubmed-ENS-20-5}}, a genetic NER model\footnote{\url{https://huggingface.co/alvaroalon2/biobert_genetic_ner}}, and a general-purpose SpaCy English NER model\footnote{\url{https://huggingface.co/spacy/en_core_web_sm}}. A warm-loading mechanism initializes all models before inference, preventing redundant loading and improving stability.}

\majorrevision{\paragraph{Concept alignment} For concept alignment, \textsc{StructSense} accesses both the local concept mapping service and BioPortal~\citep{10.1093/nar/gkaf402}, configurable via environment variables. BioPortal hosts 1,549 ontologies (1,182 public, 367 private) comprising 15,293,440 classes~\citep{10.1093/nar/gkaf402}. The local service provides access to 517 ontologies comprising 3,925,124 classes and 2,683,756 synonyms. BioPortal maps concepts primarily through lexical matching, synonym expansion, and ontology-structure-based reasoning~\citep{MartinezRomero2017}, without accounting for broader textual context---a limitation the local service addresses through contextualized hybrid retrieval.}

\majorrevision{An in-memory cache with first-in, first-out~(FIFO) eviction reduces redundant API calls. However, cached data does not persist across sessions, requiring repeated calls upon restart---an identified limitation\footnote{\url{https://github.com/sensein/structsense/issues/49}} targeted for future work. For BioPortal access, the implementation includes throttling management to handle rate limits gracefully.}

\majorrevision{\paragraph{Memory}
Three memory types\footnote{\url{https://docs.crewai.com/en/concepts/memory}} are enabled: contextual, entity, and long-term memory. These are shared across agents, allowing them to retain relevant context, previously identified entities, and knowledge acquired during execution.}

\subsection{Local Concept Mapping Service}
\label{sec:localconceptmapping}

\majorrevision{The local concept mapping service is implemented in Python using FastAPI\footnote{\url{https://fastapi.tiangolo.com}} as a standalone API service. It implements the retrieval and re-ranking pipeline described in \S\ref{sec:localconceptmappingservice}. Ontologies are retrieved from BioPortal and stored in a local SQLite database, which serves as the source of truth.}

\majorrevision{\paragraph{Search artifacts} Three artifacts are generated from the SQLite database: BM25 indexes, dense vector embeddings, and a lightweight concept cache (pickle file) that maps integer document indices to concept metadata at query time.}

\majorrevision{\paragraph{Startup strategy} A three-tier strategy avoids unnecessary rebuilding while keeping artifacts consistent with the database. At startup, the service compares the current database concept count against artifact metadata. In Tier~1, if all artifacts are valid, they are loaded directly. In Tier~2, if only the concept cache is stale, it is rebuilt from the database. In Tier~3, all artifacts are rebuilt when the ontology data changes (e.g., new ontologies are added).}

\majorrevision{\paragraph{Retrieval backends}The service supports four retrieval backends: BM25S~\citep{lu2024bm25sordersmagnitudefaster} for sparse retrieval, FAISS\footnote{\url{https://faiss.ai/index.html}} using \texttt{IndexFlatIP} for exact cosine similarity, ChromaDB\footnote{\url{https://docs.trychroma.com/docs/overview/introduction}} using HNSW indexing for approximate nearest neighbor search, and an in-memory NumPy-based backend that computes similarity via dot product over the embedding matrix. The default configuration uses FAISS with a hybrid retrieval strategy combining BM25 and dense vector search at a 30:70 ratio.}

\majorrevision{\paragraph{Re-ranking configuration} For re-ranking, the service uses \textit{jina-colbert-v2}~\citep{xiao-etal-2024-jina} as the late-interaction model. The default configuration is \textit{dual\_late}, combining late-interaction scoring (weight 0.3, normalized to 0.6) with the biomedical re-ranker (weight 0.2, normalized to 0.4). For \textsc{StructSense} integration, \textit{dual\_late} is preferred over LLM-based re-ranking because local models offer lower latency.}

\majorrevision{\paragraph{API endpoints}The service exposes three endpoints: \textit{map/concept} for single-concept mapping, \textit{map/batch} for batch mapping (up to 4,000 concepts per request), and \textit{map/search} for general search.}

\section{Experiments and Evaluation}\label{sec:experiment}

\subsection{Case Studies}\label{sec:casestudies}

\majorrevision{\paragraph{Case Study I: Neuroscience domain}\label{sec:case_studyneuroscience} We selected the neuroscience domain for its inherent complexity and interdisciplinary nature, and because it serves the requirements of three ongoing projects: Brain Behavior Quantification and Synchronization (BBQS)\footnote{\url{https://brain-bbqs.org}}, ReproNim\footnote{\url{https://www.repronim.org/index.html}}, and BICAN\footnote{\url{https://www.portal.brain-bican.org}}. Three tasks were evaluated:}

\majorrevision{\textit{Task~1: Schema-based extraction.} This task evaluates \textsc{StructSense}'s ability to convert semi-structured assessment instruments (typically PDFs) into machine-readable metadata aligned with ReproSchema~\citep{Chen2025ReproSchema}. The system must extract multi-level metadata including instrument descriptors, item properties (prompts, response options, scoring rules), and behavioral logic (e.g., conditional flows).}

\majorrevision{\textit{Task~2: Resource extraction.} This task extracts structured metadata from scientific papers, including resource type (e.g., model, dataset, tool), category (e.g., pose estimation), target entity (e.g., human, animal), description, resource name, and access URL.}

\majorrevision{\textit{Task~3: NER extraction.} This task extracts named entities---brain regions, cognitive constructs, experimental methods---from full-text neuroscience articles. It was selected because (i)~neuroscience terminology requires nuanced semantic understanding and (ii)~curated NER datasets remain scarce in this domain.}

\majorrevision{\paragraph{Case Study II: Biomedical domain}\label{sec:case_studybiomedicine} We selected the biomedical domain for its strong connection to neuroscience and the availability of established benchmarks. We evaluated NER on the NCBI Disease~\citep{DOGAN20141} dataset and the S800 Species~\citep{10.1371journal.pone.0065390} test set, sourced from BioNLP-Corpus\footnote{\url{https://github.com/bionlp-hzau/BioNLP-Corpus/tree/master}}. Our goal was not only to recover all annotated entities but also to identify additional relevant neuroscientific entities omitted by task-specific annotation guidelines. For example, in the sentence \textit{``Conversely, BRCA1 expression was reduced or undetectable in the majority of high-grade ductal carcinomas...''}, the gold annotation includes only ``ductal carcinomas'' because the benchmark is restricted to disease entities. However, our objective is broader: we aim to identify all relevant neuroscientific entities present in the text, including ``BRCA1''~(gene), which appears in the sentence but is excluded from the benchmark annotations because it falls outside their task-specific scope.}

\majorrevision{\paragraph{Case Study III: Local concept mapping}
\label{sec:caselocalconceptmappingeval} We independently evaluated the local concept mapping service to assess its retrieval accuracy as a standalone tool. The test dataset comprised 1,500 unique concepts sampled from 94 ontologies, with five query formulations per concept (Table~\ref{tab:golden_set_querytype_breakdown}). Table~\ref{tab:golden_ontology_set_properties} shows the top 10 ontologies represented. This design tested diverse, realistic query scenarios including exact-match, synonym, and alternative-label queries with and without context.}

\begin{table*}[!h]
\centering
\caption{Distribution of the test dataset for the local 
concept mapping service (1,500 unique concepts).}
\label{tab:golden_set_querytype_breakdown}
\setlength{\tabcolsep}{8pt}
\begin{tabular*}{\textwidth}{@{\extracolsep{\fill}}lrr@{}}
\toprule
\textbf{Query Type} & \textbf{Total} & \textbf{Percentage} \\
\midrule
Exact match (no context) & 623 & 41.5\% \\
Exact match (definition context) & 375 & 25.0\% \\
Synonym query (no context) & 68 & 4.5\% \\
Synonym query (other synonyms as context) & 284 & 18.9\% \\
Alt-label query (definition context) & 150 & 10.0\% \\
\midrule
\textbf{Total} & \textbf{1,500} & \textbf{100\%} \\
\bottomrule
\end{tabular*}
\smallskip
\begin{minipage}{\textwidth}
\footnotesize
\textbf{Exact match (no context)}: preferred label as query, no context. \textbf{Exact match (definition context)}: preferred label with ontology definition as context. \textbf{Synonym (no context)}: known synonym as query. \textbf{Synonym (other synonyms as context)}: synonym as query with other synonyms as context. \textbf{Alt-label (definition context)}: alternative label as query with definition as context. In all cases, the expected output is the canonical preferred label.
\end{minipage}
\end{table*}

\begin{table}[!h]
\centering
\caption{Top 10 of 94 ontologies in the test dataset by concept count.}
\label{tab:golden_ontology_set_properties}
\setlength{\tabcolsep}{5pt}
\begin{tabular*}{\columnwidth}{l@{\extracolsep{\fill}}rr}
\toprule
\textbf{Ontology} & \textbf{Count} & \textbf{Percentage} \\
\midrule
UPHENO & 284 & 21.0\% \\
RCD & 167 & 12.4\% \\
IOBC & 159 & 11.8\% \\
PR & 122 & 9.0\% \\
MEDDRA & 95 & 7.0\% \\
CHEBI & 50 & 3.7\% \\
EFO & 45 & 3.3\% \\
BERO & 44 & 3.3\% \\
HP & 40 & 3.0\% \\
NCBITAXON & 38 & 2.8\% \\
\bottomrule
\end{tabular*}
\smallskip
\begin{minipage}{\columnwidth}
\footnotesize
UPHENO = Unified Phenotype Ontology; RCD = Read Codes; 
IOBC = Interlinking Ontology for Biological Concepts; 
PR = Protein Ontology; MedDRA = Medical Dictionary for 
Regulatory Activities; ChEBI = Chemical Entities of Biological 
Interest; EFO = Experimental Factor Ontology; BERO = Biological 
and Environmental Research Ontology; HP = Human Phenotype 
Ontology; NCBITaxon = NCBI Organismal Classification.
\end{minipage}
\end{table}

\subsection{Experimental Setup}\label{sec:setup}

\majorrevision{We employed three foundation models: \textit{GPT-4o-mini} (OpenAI), \textit{Gemini 3.1 Flash Lite} (Google), and \textit{Qwen3-vl-32b-instruct}~(Alibaba Cloud), selected based on their performance\footnote{\url{https://openrouter.ai/rankings} (Accessed 18 June 2025)} and accessed via OpenRouter\footnote{\url{https://openrouter.ai}}. For embeddings, we used the open-source \textit{nomic-embed-text} model, run locally via Ollama\footnote{\url{https://ollama.com}}. All models were tested with consistent prompts, tools, and evaluation conditions. Unless otherwise noted, results are reported under the no-human-in-the-loop (NHIL) condition.}

\subsection{Evaluation Criteria}\label{sec:eval}

\majorrevision{All evaluation is fully automated, deterministic, and reproducible. Each task is evaluated by a standalone Python script using only the standard library, with no external dependencies. All scripts and result files are released alongside the codebase. It is important to note that these automated evaluation scripts are informed by domain expertise, particularly in neuroscience, and are therefore not arbitrary.}

\majorrevision{\paragraph{Task 1: Schema extraction} Ground truth is hardcoded from the source PDF of the \textit{Mood and Feelings Questionnaire: Long Version (MFQ-LV)}, a 33-item self-report depression measure scored on a 3-point Likert scale. Each model's output is scored across seven weighted dimensions: \textit{item completeness} (20\%), \textit{question text} (25\%, fuzzy matching at 85\% threshold), \textit{response options} (15\%), \textit{scoring} (15\%), \textit{activity metadata} (10\%), \textit{item order} (10\%), and \textit{input type} (5\%). Dimension scores are combined into a single weighted composite.}

\majorrevision{\paragraph{Task 2: Resource extraction} Ground truth was constructed from cross-model consensus and manual review of two papers (ViTPose and DeepLabCut). Each result is scored across three weighted dimensions: \textit{primary fields} (40\%, evaluating name, type, category, target, specific target, and URL via fuzzy and exact matching), \textit{mentions recall} (40\%, measuring recall of required and optional related entities), and \textit{ontology mapping quality} (20\%, using heuristic detection to flag cross-domain misalignments).}

\majorrevision{\paragraph{Task 3: Neuroscience NER} No gold-standard labels exist for arbitrary neuroscience papers. We construct an automated heuristic dictionary from cross-model consensus: over 200 entity-to-label mappings where at least 70\% of models agree. The pipeline proceeds in three stages: (1)~\textit{source filtering}---entities extracted exclusively by \texttt{en\_core\_web\_sm} are removed due to high noise on biomedical text; (2)~\textit{junk exclusion}---single characters, Greek letters, pure numerics, references, stopwords, URLs, and generic terms are removed; and (3)~\textit{label scoring}---each remaining entity is evaluated through a five-step cascade: multi-label check, label blacklist (clinical NER categories misapplied to neuroscience), exact dictionary match, keyword-based rules, and trusted canonical label fallback. Labels are canonicalized before comparison (e.g., \texttt{GENE\_OR\_GENE\_PRODUCT} $\to$ \texttt{GENE}, \texttt{TECHNIQUE} $\to$ \texttt{METHOD}).}

\majorrevision{\paragraph{Task 4: Biomedical NER} We evaluate entity extraction recall on two biomedical NER benchmarks: \textit{NCBI Disease} (960 mentions, 403 unique, 940 sentences) and \textit{S800 Species} (767 mentions, 370 unique, 1,630 sentences).}

\majorrevision{Because \textsc{StructSense} extracts entities across all semantic categories simultaneously, computing precision against a single-category benchmark on the full output would be misleading---a gene correctly extracted from a disease-focused paper is not a false positive. We therefore evaluate two distinct capabilities: \textit{entity detection} (did the system find the text?) and \textit{entity classification} (did it assign the correct type?).}

\majorrevision{Entity matching applies a three-tier cascade: (i)~\textit{exact match} after normalization (lowercasing, collapsing hyphens and whitespace); (ii)~\textit{span overlap} ($\geq$50\% character overlap with ground truth~(GT)); and (iii)~\textit{substring containment}. We report \textit{strict recall} (exact match only) and \textit{relaxed recall} (any tier). Recall is computed under two conditions: \textit{extraction recall}, matching against all entities regardless of label, and \textit{labeled recall}, restricted to entities whose label belongs to the target category. We additionally report \textit{precision} on the label-filtered subset: the fraction of target-labeled entities matching a ground truth mention.}

\majorrevision{\begin{align}
\text{Strict Recall} &= \frac{\text{Exact matches}}
{\text{Total GT mentions}} \\
\text{Relaxed Recall} &= \frac{\text{Any-tier matches}}
{\text{Total GT mentions}}
\end{align}}

\majorrevision{\paragraph{Task 5: Local concept mapping} Performance is measured using Hits@$K$~(Equation~\ref{eq:hitratio_at_k}) and MRR@$K$~(Equation~\ref{eq:mrr_at_k}), reported with 95\% confidence intervals. MRR@$K$ complements Hits@$K$ by capturing ranking position: a correct match at rank~3 contributes equally to Hits@$K$ (for $K \geq 3$) as one at rank~1, but receives a lower MRR@$K$ score. We evaluate under two settings: \textit{strict} (exact match to acceptable labels) and \textit{semantic} (cosine similarity $\geq$94\% to acceptable labels). Across all evaluations, we used default weights of 0.5, 0.3, and 0.2 for the LLM, late-interaction, and biomedical re-ranker components.}

\majorrevision{\begin{equation}
\text{Hits@}K = \frac{1}{N} \sum_{i=1}^{N} 
\mathbf{1}(\text{rank}_i \leq K)
\label{eq:hitratio_at_k}
\end{equation}}

\majorrevision{\begin{equation}
\text{MRR@}K = \frac{1}{N} \sum_{i=1}^{N}
\begin{cases}
\dfrac{1}{\text{rank}_i}, & \text{if } \text{rank}_i \leq K \\
0, & \text{otherwise}
\end{cases}
\label{eq:mrr_at_k}
\end{equation}}

\majorrevision{\noindent where $N$ is the total number of queries and $\text{rank}_i$ is the rank of the first correct result for query $i$.}

\section{Results}\label{sec:results}

All results are reported under the NHIL condition unless otherwise noted.

\subsection{Task 1: Schema Extraction}
\label{sec:results_schema}

\majorrevision{\begin{table*}[ht]
\centering
\caption{Schema extraction performance (Recall) on the MFQ-LV 
instrument.}
\label{tab:reproschema-results}
\resizebox{\linewidth}{!}{%
\begin{tabular}{lcccccccc}
\toprule
\textbf{Model} & \textbf{Items} & \textbf{Text} & 
\textbf{Response} & \textbf{Scoring} & \textbf{Metadata} & 
\textbf{Order} & \textbf{Input Type} & \textbf{Overall} \\
\midrule
GPT-4o-mini & 100\% & 100\% & 100\% & 100\% & 100\% & 100\% 
& 100\% & \textbf{100.0\%} \\
Gemini 3.1 Flash Lite & 100\% & 100\% & 100\% & 50\% & 88\% 
& 100\% & 100\% & \textbf{91.2\%} \\
Qwen & --- & --- & --- & --- & --- & --- & --- & 
\textbf{0.0\%} \\
\bottomrule
\end{tabular}
}
\end{table*}}

\majorrevision{Table~\ref{tab:reproschema-results} presents schema extraction results. GPT-4o-mini achieves a perfect composite score across all seven dimensions. Gemini extracts all 33 items with correct text, response options, and ordering, but omits per-item scoring maps (providing only the computed total formula), reducing its scoring dimension to 50\% and overall composite to 91.2\%. Qwen fails to produce output due to a timeout. Under HIL, Gemini scored 90.6\%, marginally below its NHIL baseline, as feedback slightly degraded metadata keyword coverage.}

\subsection{Task 2: Resource Extraction}
\label{sec:results_resource}

\begin{table*}[ht]
\centering
\caption{Resource extraction performance (Recall) across papers and 
models.}
\label{tab:resource-extraction-results}
\resizebox{\linewidth}{!}{%
\begin{tabular}{llcccc}
\toprule
\textbf{Paper} & \textbf{Model} & \textbf{Primary Fields} & 
\textbf{Mentions} & \textbf{Ontology Quality} & 
\textbf{Overall} \\
\midrule
DeepLabCut & Gemini 3.1 Flash Lite & 85\% & 100\% & 96\% & 
\textbf{93.2\%} \\
DeepLabCut & GPT-4o-mini & 100\% & 75\% & 94\% & 
\textbf{88.9\%} \\
DeepLabCut & Qwen & 75\% & 100\% & 98\% & 
\textbf{89.4\%} \\
\midrule
ViTPose & Gemini 3.1 Flash Lite & 94\% & 100\% & 41\% & 
\textbf{85.9\%} \\
ViTPose & GPT-4o-mini & 94\% & 75\% & 97\% & 
\textbf{87.2\%} \\
ViTPose & Qwen & 72\% & 96\% & 97\% & 
\textbf{86.6\%} \\
\bottomrule
\end{tabular}
}
\end{table*}

\majorrevision{Table~\ref{tab:resource-extraction-results} presents resource extraction results. All models correctly identify primary fields (name, type, category, target) in nearly all cases. Gemini and Qwen achieve higher mention recall (96--100\%) than GPT-4o-mini (75\%), which misses tools such as \texttt{idtracker.ai} and \texttt{mmpose}. Ontology mapping quality is the weakest and most variable dimension: Gemini on ViTPose scores only 41\% because the ontology lookup service maps ``COCO'' to the plant species \textit{Magnolia coco} and model names to pharmaceutical products---a limitation of the service rather than the extraction agent. Under HIL, Gemini on ViTPose dropped to 75.8\% due to a lost tool mention during feedback processing.}

\subsection{Task 3: Neuroscience NER}
\label{sec:results_neuro_ner}

\begin{table*}[ht]
\centering
\caption{NER label correctness by model (aggregated across 
3 papers).}
\label{tab:ner-accuracy}
\begin{tabular}{lrrrc}
\toprule
\textbf{Model} & \textbf{Total} & \textbf{Evaluated} & 
\textbf{Correct} & \textbf{Accuracy} \\
\midrule
Qwen & 2,358 & 1,310 & 898 & \textbf{75.4\%} \\
Gemini 3.1 Flash Lite & 2,835 & 2,527 & 1,754 & 
\textbf{69.0\%} \\
GPT-4o-mini & 3,689 & 2,994 & 1,773 & \textbf{58.6\%} \\
\bottomrule
\end{tabular}
\end{table*}

\majorrevision{Table~\ref{tab:ner-accuracy} presents aggregate NER label correctness. The system extracted 8,882 entities across 9 result files; 6,831 remained after source filtering and junk exclusion. Of these, 4,425 (64.8\%) received correct labels, 2,008 (29.4\%) incorrect labels, and 398 (5.8\%) had no heuristic match. Excluding unknowns, overall accuracy is 68.8\%.}

\majorrevision{Qwen achieves the highest accuracy (75.4\%) despite extracting the fewest entities, suggesting a conservative but precise strategy. GPT-4o-mini extracts the most entities but scores the lowest accuracy (58.6\%). Performance varies across papers: 65.1\% (\textit{Discovery of Optimal Cell Type Classification}), 62.1\% (\textit{Latent Circuit Inference}), and 78.9\% (\textit{Multiscale Spatial Transcriptomic}). Qwen produced no output for the third paper.}

\majorrevision{Table~\ref{tab:ner-per-label} breaks down accuracy by canonical entity type. Categories with a well-defined domain semantics---Cell Type (99\%), Species (99\%), Metric (99\%), Gene (94\%), Brain Region (94\%)---achieve near-perfect accuracy. Broader categories degrade: Biological Process (83\%), Concept (87\%), Organization (54\%), Product (51\%). Two systematic failure modes account for 1,434 entities at 0\%: clinical NER labels misapplied by \texttt{d4data/biomedical-ner-all} (705 entities) and uncategorized catch-all labels such as \texttt{ENTITY}, \texttt{MISC}, and \texttt{OTHER} (729 entities).}

\begin{table*}[ht]
\centering
\caption{NER label correctness by entity type}
\label{tab:ner-per-label}
\begin{tabular}{p{2.5cm}rrrr}
\toprule
\textbf{Entity Type} & \textbf{Gemini} & \textbf{GPT-4o-mini} 
& \textbf{Qwen} & \textbf{All} \\
\midrule
CONCEPT & 81\% (164/203) & 95\% (274/288) & 84\% (283/337) & 
\textbf{87\%} (721/828) \\
BRAIN\_REGION & 96\% (371/386) & 93\% (274/294) & 89\% (51/57) 
& \textbf{94\%} (696/737) \\
UNCATEGORIZED & 0\% (0/2) & 0\% (0/671) & 0\% (0/56) & 
\textbf{0\%} (0/729) \\
CLINICAL\_NER & 0\% (0/578) & 0\% (0/126) & 0\% (0/1) & 
\textbf{0\%} (0/705) \\
CELL\_TYPE & 100\% (294/294) & 99\% (265/269) & 98\% (59/60) & 
\textbf{99\%} (618/623) \\
METHOD & 92\% (208/226) & 87\% (211/243) & 92\% (85/92) & 
\textbf{90\%} (504/561) \\
GENE & 100\% (217/218) & 96\% (183/190) & 75\% (69/92) & 
\textbf{94\%} (469/500) \\
BIOLOGICAL PROCESS & 66\% (23/35) & 87\% (126/145) & 
84\% (156/186) & \textbf{83\%} (305/366) \\
ORGANIZATION & 73\% (55/75) & 42\% (69/163) & 66\% (25/38) & 
\textbf{54\%} (149/276) \\
METRIC & 100\% (59/59) & 99\% (101/102) & 99\% (69/70) & 
\textbf{99\%} (229/231) \\
MODEL & 99\% (118/119) & 94\% (58/62) & 94\% (30/32) & 
\textbf{97\%} (206/213) \\
PRODUCT & --- & 63\% (12/19) & 49\% (70/143) & 
\textbf{51\%} (82/162) \\
CHEMICAL & 93\% (54/58) & 98\% (41/42) & 84\% (43/51) & 
\textbf{91\%} (138/151) \\
SPECIES & 100\% (65/65) & 100\% (39/39) & 89\% (8/9) & 
\textbf{99\%} (112/113) \\
SOFTWARE & 94\% (50/53) & 100\% (24/24) & 91\% (20/22) & 
\textbf{95\%} (94/99) \\
PARAMETER & 100\% (21/21) & 94\% (15/16) & 100\% (6/6) & 
\textbf{98\%} (42/43) \\
DATASET & 88\% (15/17) & 95\% (18/19) & 100\% (5/5) & 
\textbf{93\%} (38/41) \\
MOLECULE & 100\% (25/25) & 100\% (13/13) & 50\% (1/2) & 
\textbf{98\%} (39/40) \\
DATA\_TYPE & 100\% (5/5) & 100\% (16/16) & 93\% (14/15) & 
\textbf{97\%} (35/36) \\
STIMULUS & 100\% (22/22) & 100\% (8/8) & 100\% (4/4) & 
\textbf{100\%} (34/34) \\
DISEASE & 100\% (6/6) & 92\% (11/12) & 80\% (4/5) & 
\textbf{91\%} (21/23) \\
CELLULAR COMPONENT & --- & 75\% (6/8) & 78\% (7/9) & 
\textbf{76\%} (13/17) \\
VARIABLE & 100\% (6/6) & 100\% (8/8) & 100\% (2/2) & 
\textbf{100\%} (16/16) \\
BEHAVIOR & --- & 91\% (10/11) & 100\% (4/4) & 
\textbf{93\%} (14/15) \\
MATERIAL & 100\% (3/3) & 100\% (11/11) & --- & 
\textbf{100\%} (14/14) \\
PERSON & 100\% (3/3) & 100\% (9/9) & 100\% (1/1) & 
\textbf{100\%} (13/13) \\
DEVICE & --- & 100\% (9/9) & 100\% (1/1) & 
\textbf{100\%} (10/10) \\
\midrule
Other (rare) & 8\% (4/48) & 2\% (4/177) & 0\% (0/10) & 
\textbf{3\%} (8/235) \\
\bottomrule
\end{tabular}
\end{table*}

\subsection{Biomedical NER Benchmarks}
\label{sec:results_biomed_ner}

\begin{table*}[ht]
\centering
\caption{Benchmark NER Performance on NCBI Disease and S800 Species}
\label{tab:benchmark-ner}
\begin{tabular}{p{2cm}p{2cm}ccccccc}
\toprule
 & & \multicolumn{2}{c}{\textbf{Extraction Recall}} & \multicolumn{2}{c}{\textbf{Labeled Recall}} & & \\
\cmidrule(lr){3-4} \cmidrule(lr){5-6}
\textbf{Dataset} & \textbf{Model} & \textbf{Strict} & \textbf{Relaxed} & \textbf{Strict} & \textbf{Relaxed} & \textbf{Precision} & \textbf{Extra} \\
\midrule
NCBI Disease & Gemini 3.1 Flash Lite & 83.1\% & 96.1\% & 77.6\% & 90.6\% & 86.2\% & 1,049 \\
NCBI Disease & GPT-4o-mini           & 80.9\% & 94.5\% & 73.5\% & 88.0\% & 87.1\% & 1,192 \\
NCBI Disease & Qwen                  & 84.1\% & 97.0\% & 78.8\% & 92.7\% & 74.2\% & 1,504 \\
\midrule
S800 Species & Gemini 3.1 Flash Lite & 85.8\% & 96.6\% & 75.7\% & 89.7\% & 60.1\% & 2,719 \\
S800 Species & GPT-4o-mini           & 62.5\% & 81.0\% & 50.5\% & 65.1\% & 52.4\% & 3,223 \\
S800 Species & Qwen                  & 70.0\% & 90.6\% & 39.5\% & 56.1\% & 47.6\% & 3,622 \\
\bottomrule
\end{tabular}
\end{table*}

\majorrevision{Table~\ref{tab:benchmark-ner} presents the benchmark NER recall results on two standard biomedical datasets.}

\majorrevision{Extraction recall (label-agnostic) exceeds 90\% for most configurations, reaching 97.0\%~(Qwen on NCBI Disease). Strict recall ranges from 62.5\% to 85.8\%}

\majorrevision{Labeled recall, requiring both correct extraction and labeling, remains high on NCBI Disease (88.0--92.7\%, a 4--7 point drop from extraction recall). On S800 Species, the gap widens: Qwen drops from 90.6\% to 56.1\% and GPT-4o-mini from 81.0\% to 65.1\%, as many species entities are mislabeled as \texttt{GENE}, \texttt{BIOLOGICAL\_PROCESS}, or \texttt{PROTEIN}. Gemini maintains the highest labeled recall and the smallest gap on both datasets.}

\majorrevision{Precision is highest on NCBI Disease~(GPT-4o-mini 87.1\%, Gemini 86.2\%) while on S800 Species, precision drops to 47.6--60.1\%}

\majorrevision{All models extract 1,000--3,600 entities beyond gold annotations, including genes, proteins, chemicals, and biological processes. Under HIL, relaxed recall improves marginally (e.g., Qwen on NCBI: 97.5\% vs.\ 97.0\%) while strict recall slightly decreases.}

\majorrevision{All models extract 1,000--3,600 entities beyond gold annotations, including genes, proteins, chemicals, and biological processes. Under HIL, Qwen on NCBI Disease improves marginally (97.5\% vs.\ 97.0\% relaxed) but slightly decreases in strict recall, consistent with feedback refining entity text in ways that diverge from ground truth formatting.}

\subsection{Local Concept Mapping}
\label{sec:results_concept_mapping}

Tables~\ref{tab:all_results_strict} 
and~\ref{tab:all_results_semantic} present retrieval 
performance under strict and semantic matching, respectively.

\begin{table*}[!h]
\centering 
\caption{Retrieval performance under strict matching.}
\label{tab:all_results_strict}
\scriptsize
\setlength{\tabcolsep}{4pt}
\resizebox{\linewidth}{!}{%
\begin{tabular}{llccc}
\toprule
\textbf{Aggregate} & \textbf{Re-ranker} & \textbf{@1} & 
\textbf{@3} & \textbf{@5} \\
 & & \textbf{Hits / MRR} & \textbf{Hits / MRR} & 
\textbf{Hits / MRR} \\
\midrule
Micro & single\_late\_interaction & 77.6\% [76.4, 78.8] / 
77.6\% [76.4, 78.8] & 85.2\% [84.1, 86.2] / 81.0\% [80.0, 
82.1] & 87.7\% [86.7, 88.7] / 81.6\% [80.6, 82.7] \\
 & single\_biomedical & 62.9\% [61.5, 64.3] / 62.9\% [61.5, 
64.3] & 80.8\% [79.6, 81.9] / 70.9\% [69.7, 72.1] & 86.4\% 
[85.4, 87.4] / 72.2\% [71.1, 73.3] \\
 & single\_llm & \textbf{81.8\%} [80.6, 82.9] / 
\textbf{81.8\%} [80.7, 82.9] & \textbf{89.0\%} [88.1, 89.9] / 
\textbf{85.1\%} [84.1, 86.0] & \textbf{90.8\%} [90.0, 91.7] / 
\textbf{85.5\%} [84.5, 86.4] \\
\cmidrule(lr){2-5}
 & dual\_llm\_late & \underline{79.6\%} [78.4, 80.7] / 
\underline{79.6\%} [78.4, 80.7] & 85.6\% [84.5, 86.6] / 
\underline{82.2\%} [81.1, 83.3] & 88.1\% [87.1, 89.0] / 
\underline{82.8\%} [81.7, 83.8] \\
 & dual\_llm\_biomedical & 76.4\% [75.2, 77.7] / 76.4\% 
[75.2, 77.7] & \underline{87.0\%} [86.0, 87.9] / 81.2\% 
[80.1, 82.2] & \underline{89.8\%} [88.9, 90.7] / 81.8\% 
[80.8, 82.8] \\
 & dual\_late & 71.3\% [70.0, 72.6] / 71.3\% [70.0, 72.7] & 
82.9\% [81.7, 83.9] / 76.5\% [75.4, 77.7] & 86.8\% [85.8, 
87.8] / 77.4\% [76.3, 78.5] \\
\cmidrule(lr){2-5}
 & ensemble & 78.3\% [77.1, 79.5] / 78.3\% [77.1, 79.5] & 
85.3\% [84.2, 86.3] / 81.4\% [80.4, 82.5] & 88.0\% [87.0, 
88.9] / 82.0\% [81.0, 83.1] \\
\midrule
Macro & single\_late\_interaction & 77.6\% / 77.6\% & 85.2\% / 
81.0\% & 87.7\% / 81.6\% \\
 & single\_biomedical & 62.9\% / 62.9\% & 80.8\% / 70.9\% & 
86.4\% / 72.2\% \\
 & single\_llm & \textbf{81.8\%} / \textbf{81.8\%} & 
\textbf{89.0\%} / \textbf{85.1\%} & \textbf{90.8\%} / 
\textbf{85.5\%} \\
\cmidrule(lr){2-5}
 & dual\_llm\_late & \underline{79.6\%} / \underline{79.6\%} & 
85.6\% / \underline{82.2\%} & 88.1\% / \underline{82.8\%} \\
 & dual\_llm\_biomedical & 76.4\% / 76.4\% & 
\underline{87.0\%} / 81.2\% & \underline{89.8\%} / 81.8\% \\
 & dual\_late & 71.3\% / 71.3\% & 82.9\% / 76.5\% & 86.8\% / 
77.4\% \\
\cmidrule(lr){2-5}
 & ensemble & 78.3\% / 78.3\% & 85.3\% / 81.4\% & 88.0\% / 
82.0\% \\
\bottomrule
\end{tabular}%
}
\par\smallskip
\begin{minipage}{\linewidth}
\footnotesize
Hits@k / MRR@k. Micro: all query--endpoint pairs; macro: 
endpoint-level averages. 95\% confidence intervals shown. 
\textbf{Bold} = best; \underline{underlined} = second-best.
\end{minipage}
\end{table*}

\begin{table*}[t]
\centering
\caption{Retrieval performance under semantic matching.}
\label{tab:all_results_semantic}
\scriptsize
\setlength{\tabcolsep}{4pt}
\resizebox{\linewidth}{!}{%
\begin{tabular}{llccc}
\toprule
\textbf{Aggregate} & \textbf{Re-ranker} & \textbf{@1} & 
\textbf{@3} & \textbf{@5} \\
 & & \textbf{Hits / MRR} & \textbf{Hits / MRR} & 
\textbf{Hits / MRR} \\
\midrule
Micro & single\_late\_interaction & 81.8\% [80.6, 82.9] / 
81.8\% [80.7, 82.9] & 87.5\% [86.5, 88.4] / 84.3\% [83.3, 
85.3] & 89.7\% [88.7, 90.5] / 84.8\% [83.9, 85.8] \\
 & single\_biomedical & 68.5\% [67.2, 69.9] / 68.5\% [67.2, 
69.9] & 84.2\% [83.1, 85.2] / 75.5\% [74.4, 76.6] & 88.9\% 
[87.9, 89.8] / 76.6\% [75.5, 77.7] \\
 & single\_llm & \textbf{85.8\%} [84.7, 86.8] / 
\textbf{85.8\%} [84.8, 86.8] & \textbf{91.0\%} [90.2, 91.8] / 
\textbf{88.1\%} [87.2, 89.0] & \textbf{92.4\%} [91.5, 93.1] / 
\textbf{88.4\%} [87.5, 89.3] \\
\cmidrule(lr){2-5}
 & dual\_llm\_late & \underline{82.8\%} [81.7, 83.9] / 
\underline{82.8\%} [81.7, 83.9] & 87.9\% [86.9, 88.8] / 
\underline{85.0\%} [84.0, 86.0] & 90.1\% [89.2, 90.9] / 
\underline{85.5\%} [84.5, 86.5] \\
 & dual\_llm\_biomedical & 80.6\% [79.4, 81.7] / 80.6\% 
[79.4, 81.7] & \underline{89.2\%} [88.3, 90.1] / 84.5\% 
[83.5, 85.5] & \underline{91.3\%} [90.5, 92.1] / 85.0\% 
[84.0, 85.9] \\
 & dual\_late & 75.6\% [74.3, 76.8] / 75.6\% [74.3, 76.9] & 
85.5\% [84.4, 86.5] / 80.0\% [79.0, 81.1] & 88.8\% [87.8, 
89.7] / 80.8\% [79.8, 81.8] \\
\cmidrule(lr){2-5}
 & ensemble & 81.7\% [80.5, 82.8] / 81.7\% [80.6, 82.8] & 
87.9\% [86.9, 88.8] / 84.4\% [83.4, 85.4] & 90.1\% [89.2, 
90.9] / 84.9\% [83.9, 85.9] \\
\midrule
Macro & single\_late\_interaction & 81.8\% / 81.8\% & 87.5\% / 
84.3\% & 89.7\% / 84.8\% \\
 & single\_biomedical & 68.5\% / 68.5\% & 84.2\% / 75.5\% & 
88.9\% / 76.6\% \\
 & single\_llm & \textbf{85.8\%} / \textbf{85.8\%} & 
\textbf{91.0\%} / \textbf{88.1\%} & \textbf{92.4\%} / 
\textbf{88.4\%} \\
\cmidrule(lr){2-5}
 & dual\_llm\_late & \underline{82.8\%} / \underline{82.8\%} & 
87.9\% / \underline{85.0\%} & 90.1\% / \underline{85.5\%} \\
 & dual\_llm\_biomedical & 80.6\% / 80.6\% & 
\underline{89.2\%} / 84.5\% & \underline{91.3\%} / 85.0\% \\
 & dual\_late & 75.6\% / 75.6\% & 85.5\% / 80.0\% & 88.8\% / 
80.8\% \\
\cmidrule(lr){2-5}
 & ensemble & 81.7\% / 81.7\% & 87.9\% / 84.4\% & 90.1\% / 
84.9\% \\
\bottomrule
\end{tabular}%
}
\par\smallskip
\begin{minipage}{\linewidth}
\footnotesize
Hits@k / MRR@k. Micro: all query--endpoint pairs; macro: 
endpoint-level averages. 95\% confidence intervals shown. 
\textbf{Bold} = best; \underline{underlined} = second-best.
\end{minipage}
\end{table*}

\majorrevision{Under strict matching (Table~\ref{tab:all_results_strict}), \textit{single\_llm} achieves the best performance at all cutoffs: Hits@1 of 81.8\%, Hits@3 of 89.0\%, and Hits@5 of 90.8\%. Among dual configurations, \textit{dual\_llm\_late} ranks second at @1 (79.6\%), while \textit{dual\_llm\_biomedical} reaches the second-highest Hits@5 (89.8\%). The fully local \textit{dual\_late} configuration remains competitive at 86.8\% Hits@5 without any LLM dependency. \textit{Single\_biomedical} shows the lowest rank-1 performance (62.9\%) but recovers to 86.4\% at Hits@5, indicating that relevant concepts are retrieved but not consistently top-ranked---likely due to its heuristic-based implementation.}

\majorrevision{Under semantic matching (Table~\ref{tab:all_results_semantic}), method rankings remain consistent with strict matching. Scores increase by 2--5 percentage points across all configurations.}

\section{Discussion}\label{sec:discussion}

\majorrevision{In this study, we evaluated \textsc{StructSense} across three case studies using three foundation models (Gemini 3.1 Flash Lite, GPT-4o-mini, and Qwen). Case Study~I tested three neuroscience-domain tasks: schema-based extraction, resource extraction, and NER from scientific literature. Case Study~II benchmarked biomedical NER on NCBI Disease and S800 Species. Case Study~III assessed the local concept mapping service independently.}

\majorrevision{\subsection{Effect of Task Structure on Extraction Quality}}

\majorrevision{Performance varied inversely with the semantic openness of each task. In Case Study~I, Task~1 (schema extraction) defined a fixed target structure---33 items, three response options, and explicit scoring rules---leaving little room for interpretation. Both GPT-4o-mini (100\%) and Gemini (91.2\%) achieved high composite scores; Gemini's only deficit was the omission of per-item scoring maps. Task~2 (resource extraction) introduced moderate variability: models had to identify resource names, types, categories, and related mentions from free-text descriptions. All models scored above 85\% overall, yet ontology mapping quality fluctuated substantially---Gemini on ViTPose dropped to 41\% due to cross-domain misalignment by the ontology lookup service (e.g., ``COCO'' mapped to the plant species \textit{Magnolia coco}). Task~3 (neuroscience NER) presented the most open-ended setting, requiring the identification and labeling of diverse entities from full-text articles without predefined entity lists. Label accuracy ranged from 58.6\% to 75.4\%, and performance varied across papers~(62.1--78.9\%), reflecting differences in domain complexity and entity ambiguity. This gradient suggests that as extraction tasks move from deterministic to interpretive, the demand for external knowledge grounding and human oversight increases.}

\majorrevision{For Case Study~II, though the task nature is similar to Task~3 (neuroscience NER), all models extract 1,000--3,600 entities beyond gold annotations, demonstrating \textsc{StructSense}'s broader extraction capability as a multi-NER aggregation system. Precision is highest on NCBI Disease but drops to 47.6--60.1\%, reflecting difficulty distinguishing species from genes and proteins in abbreviated taxonomic nomenclature. The performance difference could be attributed to the fact that the benchmark dataset is curated specifically for the NER task and inherently constrained, whereas Task~3 requires extraction from PDFs, introducing an additional preprocessing step that may affect data quality.}

\majorrevision{\subsection{Impact of Local Concept Mapping Service}}

\majorrevision{The concept mapping service plays a critical role in disambiguation. As illustrated in Table~\ref{tab:cortex-ambiguity}, the polysemous term ``cortex'' maps to UBERON\_0000956 in neuroscience but to PO\_0005708 in plant biology. Without ontological grounding, an LLM may select the wrong sense. The alignment agent, with access to domain-specific ontologies, resolves such ambiguities by linking each mention to the appropriate ontology term.}

\majorrevision{The local concept mapping service demonstrated consistent retrieval performance under both strict and semantic matching at micro and macro levels. Among re-ranking configurations, \textit{single\_llm} achieved the highest Hits@1 (81.8\% strict, 85.8\% semantic), confirming the effectiveness of LLM-based re-ranking for concept selection. However, non-LLM configurations remained competitive: \textit{dual\_late}, which combines late-interaction and biomedical re-rankers without any LLM, reached 86.8\% Hits@5 under strict matching. This result suggests that concept mapping can be supported effectively by local open-source models, reducing dependence on proprietary APIs.}

\majorrevision{Semantic matching consistently improved scores by 2--5 percentage points over strict matching across all configurations, indicating that many near-miss predictions remained semantically close to acceptable labels. The largest gains appeared for \textit{single\_biomedical} (+5.6 points at Hits@1), which retrieves conceptually related candidates even when exact lexical agreement fails.}

\majorrevision{\subsection{Role of Human-in-the-Loop Feedback}}

\majorrevision{HIL feedback produced mixed effects across tasks. For biomedical NER, human feedback provided marginal improvement on relaxed recall (e.g., Qwen on NCBI Disease: 97.5\% vs.\ 97.0\% NHIL) but slightly decreased strict recall, consistent with feedback refining entity text in ways that diverge from ground truth formatting. For schema extraction, Gemini scored 90.6\% under HIL versus 91.2\% under NHIL, as the feedback step slightly degraded metadata keyword coverage. For resource extraction, Gemini on ViTPose dropped from 85.9\% to 75.8\% under HIL due to a lost tool mention during feedback processing.}

\majorrevision{These observations suggest that HIL is most beneficial when extraction errors stem from missing domain knowledge or ambiguous entity boundaries, rather than from formatting or surface-level text differences. Designing feedback mechanisms that selectively target high-uncertainty predictions, rather than applying uniform post-processing, represents an important direction for future work.}

\majorrevision{\subsection{Comparison with Existing Works}}

\majorrevision{\textsc{StructSense} advances the state of the art along several axes. First, it enforces explicit source grounding: all extracted entities are verified against the original text at both lexical and sentence levels, and entities failing verification are discarded. This addresses the limitation identified by~\cite{Dagdelen2024}, whose system lacks systematic grounding of generated outputs in source evidence. Second, \textsc{StructSense} systematically captures provenance information, recording the source model for each extracted entity and distinguishing whether concept mappings originate from external tools or from the LLM's parametric knowledge. Such traceability is absent in most existing LLM-based IE systems~\citep{agrawal2022largelanguagemodelsfewshot, wei2023claimdistiller}.}

\majorrevision{Third, unlike rule-based approaches~\citep{9366868, Schröder2022} that lack flexibility in handling linguistic variability, and unlike ML-based approaches~\citep{10101029, 8334468} that require domain-specific training data, \textsc{StructSense} operates in a task-agnostic manner across diverse extraction settings without fine-tuning. Its modular architecture allows seamless integration of additional domain-specific models, tools, and verification mechanisms. Fourth, the integration of ontology-guided alignment addresses the schema adherence challenge noted by Agrawal et al.~\citep{agrawal2022largelanguagemodelsfewshot}, who observed that LLMs struggle to conform to exact schemas without external tool support.}

\majorrevision{\subsection{Limitations}}

\majorrevision{Several limitations should be noted.
First, the ontology mapping service occasionally produced cross-domain misalignments~(e.g., ``COCO'' mapped to \textit{Magnolia coco}), a limitation of the external ontology lookup rather than the extraction agent itself but one that affects end-to-end output quality. It is because we have the limited ontologies, i.e., 517 ontologies compared to BioPortal's 1,549, limiting concept coverage for less common domains. Second, the neuroscience NER evaluation relied on heuristic consensus dictionaries built based on human expertise rather than purely manual annotations from domain experts, which may introduce systematic biases in accuracy estimates.}

\section{Conclusion and Future Work}\label{sec:conclusion}

\majorrevision{We introduced \textsc{StructSense}, a modular, task-agnostic multi-agent framework that integrates ontology-guided knowledge grounding, self-evaluative refinement, and human-in-the-loop validation for structured information extraction. Across three case studies, the system achieved 91--100\% on schema extraction, 86--93\% on resource extraction, 58--75\% on neuroscience NER, and $\geq$90\% relaxed recall on two biomedical NER benchmarks while extracting 1,000--3,600 entities beyond gold annotations. The local concept mapping service reached Hits@1 of 62--82\% under strict matching and 68--86\% under semantic matching. Beyond performance, \textsc{StructSense} enforces explicit source grounding, captures systematic provenance, and supports configurable concept mapping, addressing key gaps in existing IE systems. Future work should integrate domain-appropriate NER models, expand ontology coverage, develop expert-annotated neuroscience benchmarks, and extend the framework to relation extraction and cross-study knowledge synthesis.}

\bibliographystyle{abbrvnat}
\bibliography{ref}  

\section*{Acknowledgments}
This work is supported by NIH awards U24MH130918~(BICAN Knowledgebase), P41EB019936~(ReproNim), U24MH136628~(BBQS) and R24MH117295~(DANDI). We would like to acknowledge Bruke Wossenseged for his assistance during the early stages of understanding the resource extraction task and its requirements, and \majorrevision{Nader Nikbakht, Ph.D., for valuable feedback on the figures. We are especially grateful to Aditya Joshi (University of Notre Dame) for externally testing the local concept mapping service and providing suggestions for usability. We also thank Yun~(Renee) Zhang, Ph.D., and colleagues at the NIH~(National Institutes of Health) NLM~(National Library of Medicine) for their feedback during our discussions, which helped strengthen our evaluation. Finally, we acknowledge the BICAN AI Interest Group for their insightful feedback, which contributed to further improving the evaluation.}

\end{document}